\documentclass{article}

\usepackage{microtype}
\usepackage{graphicx}
\usepackage{subfigure}
\usepackage{booktabs} 
\usepackage{array}
\usepackage{amssymb}
\usepackage{amsmath}
\usepackage{tabularx}
\usepackage{lipsum}
\usepackage{placeins}
\usepackage{balance}

\usepackage{hyperref}

\usepackage[accepted]{mlsys2024}

\mlsystitlerunning{Efficient Post-training Quantization with FP8 Formats}

\begin{document}

\twocolumn[
\mlsystitle{Efficient Post-training Quantization with FP8 Formats}



\mlsyssetsymbol{equal}{*}

\begin{mlsysauthorlist}
\mlsysauthor{Haihao Shen}{inc}
\mlsysauthor{Naveen Mellempudi}{amd,equal}
\mlsysauthor{Xin He}{inc}
\mlsysauthor{Qun Gao}{inc}
\mlsysauthor{Chang Wang}{inc}
\mlsysauthor{Mengni Wang}{inc}
\end{mlsysauthorlist}

\mlsysaffiliation{inc}{Intel Corporation, Shanghai, China.} 
\mlsysaffiliation{amd}{AMD, Austin, Texas, United States} 

\mlsyscorrespondingauthor{Haihao Shen}{haihao.shen@intel.com}

\mlsyskeywords{Machine Learning, MLSys}

\vskip 0.3in

\begin{abstract}
Recent advances in deep learning methods such as LLMs and Diffusion models have created a need for improved quantization methods that can meet the computational demands of these modern architectures while maintaining accuracy. Towards this goal, we study the advantages of FP8 data formats for post-training quantization across 75 unique network architectures covering a wide range of tasks, including machine translation, language modeling, text generation, image classification, generation, and segmentation. We examine three different FP8 representations (E5M2, E4M3, and E3M4) to study the effects of varying degrees of trade-off between dynamic range and precision on model accuracy. Based on our extensive study, we developed a quantization workflow that generalizes across different network architectures. Our empirical results show that FP8 formats outperform INT8 in multiple aspects, including workload coverage (92.64\% vs. 65.87\%), model accuracy and suitability for a broader range of operations. Furthermore, our findings suggest that E4M3 is better suited for NLP models, whereas E3M4 performs marginally better than E4M3 on computer vision tasks. 
\end{abstract}
]



\printAffiliationsAndNotice{\mlsysEqualContribution} 
\section{Introduction}
\label{intro}

Quantization is the process of reducing the numeric precision of weights and activations of a neural network to lower the computation costs of inference. INT8 quantization~\cite{int8_vanhoucke, songhan_deep_compression} is the most widely-accepted choice today due to its ability to deliver high inference performance on modern deep learning hardware while maintaining reasonable model accuracy.
It has been particularly effective for computer vision tasks such as object detection and image classification, and has been widely deployed in production both at the data center scale and on resource-constrained edge devices. However, INT8 presents several challenges that arise due to its limited dynamic range. Several quantization techniques have been developed to address these challenges. For example, asymmetric quantization~\cite{Jacob_2018_CVPR, krishnamoorthi_quant, yash_asymmetric} allocates different numbers of bits for the positive and negative ranges with a non-zero offset, to better represent the distribution of the original values. Non-uniform quantization methods~\cite{log_quant_Miyashita, incremental_quant, half_wave_gaussian, piecewise_linear, additive_powerof2} attempt to assign more precision to the parts of the data that are deemed more important to reduce quantization errors. Methods that use per-group~\cite{dorefa, fgq} or per-channel \cite{Jacob_2018_CVPR, krishnamoorthi_quant} scaling extend the effective dynamic range by using independent scaling factor for each selected group of elements. The limited dynamic range of INT8 also results in poor representation of outliers that are typically found in activations. This is especially prevalent in Large Language Models (LLMs), where outliers are significantly larger when compared to the rest of the activations. Most common approach for handling outliers is to clip them using threshold values that are either obtained through calibration~\cite{clipping_resiliancy, outlier_splitting} 
or learned during training~\cite{yash_asymmetric, PACT, learned_step, lq_nets}. More recently~\cite{outlier_supression, xiao2022smoothquant} have proposed applying mathematical transformations to redistribute the magnitude of outliers between weights and activation tensors to minimize their impact. Despite these advancements, INT8 methods remain ineffective for a wide range of language modeling tasks, where the presence of LayerNorm was shown to amplify the occurrence of outliers~\cite{outlier_supression}. Therefore, a significant percentage of these workloads falls back to using higher precision to preserve model accuracy. 

\begin{table*}[t]
\caption{FP8 binary formats: The E\emph{\underline{e}}M\textit{\underline{m}} notation represents bit allocation for \emph{Exponent (e)} and \emph{Mantissa (m)} respectively. The formats support a \emph{sign-bit} and 
an implicit leading bit in the mantissa. E5M2 follows IEEE-like encoding rules, while E4M3 and E3M4 use extended encoding to reclaim $\pm$Infinity for useful encoding, a unique bit-sequence of \emph{all-ones} represents a NaN.}
\label{encodings-table}
\vskip 0.15in
\begin{center}
\begin{small}
\begin{sc}
\begin{tabular}{p{4cm}>{\centering}p{3cm}>{\centering}p{3cm}>{\centering\arraybackslash}p{3cm}}
\toprule
  & E5M2 & E4M3 & E3M4 \\
\midrule
Exponent bias (\emph{b})  &   15   & 7    & 3 \\
Max value & 57344.0& 448.0& 30.0 \\
Min value & $1.5\times10^{-5}$ & $1.9\times10^{-3}$ & $1.5\times10^{-2}$ \\
Subnormals & Yes & Yes & Yes \\
NaNs      & all & single & single \\
Infinity  & Yes & No     & No       \\
\bottomrule
\end{tabular}
\end{sc}
\end{small}
\end{center}
\vskip 0.15in
\end{table*}

This paper argues that 8-bit floating-point (FP8) formats are an efficient and more productive alternative to INT8 for deep neural network quantization. We evaluated three different representations (E5M2, E4M3, and E3M4) that offer varying degrees of trade-off between dynamic range and precision. Table 1 shows the details of the binary format and special value encoding. The study focused on the benefits of FP8 formats for post-training quantization as the preferred approach used in production. We developed quantization workflows that generalized across different network architectures, and conducted experiments on 75 networks that cover a wide range of application domains. Our results show that FP8 formats overall provide higher accuracy, better workload coverage compared to INT8 (92.64\% vs. 65.87\%) and can handle more operations such as LayerNorm and BatchNorm. The data also suggests that E4M3 is better suited for a broad range of NLP models with a coverage of 96.32\% compared to E3M4 (92.11\%), while E3M4 performs slightly better on computer vision models with 78.95\% coverage compared to E4M3 (73.68\%). Our contributions are as follows:

\begin{itemize}
\item Propose a unified and scalable FP8 quantization flow that works across application domains and different model sizes. To the best of our knowledge, our work is the first to study this problem across 200+ tasks and 75+ models demonstrating the scalability of our approach. 
\item Demonstrate the advantages of FP8 formats over INT8, in terms of workload coverage, model accuracy and suitability for a broader range of operations. Our work is also the first study to showcase accuracy-driven automatic model tuning for quantization. 
\item Suggest that E4M3 is better suited for NLP models, whereas E3M4 performs marginally better than E4M3 on computer vision tasks.
\end{itemize}

\subsection{Related Work}

There is a growing body of research is studying the use of 8-bit floating-point formats to accelerate deep learning training and inference tasks. Initial studies by \cite{wang_training_2018} and \cite{ mellempudi_mixed_2019} focused on the E5M2 format for training tasks due to its wider dynamic range which is necessary for representing gradient values. \cite{sun_hybrid_2019} subsequently proposed using a combination of two binary formats, E5M2 and E4M3, for training and extended their research to include inference tasks. They also suggested using an exponent bias to shift the numeric range of E4M3 format for handling outliers in activations. Later studies by \cite{noune_8-bit_2022} and \cite{kuzmin_fp8_2022} have extended this scope to include variable exponent bias and formats with fewer exponent bits, such as E3M4 and E2M5. More recently, \cite{micikevicius2022fp8} presented a generalized training method that employs per-tensor scaling using E5M2 and E4M3 formats. They also extended the inference studies to cover large language models such as GPT-3 (6.7B).

The rest of this paper is organized as follows. Section \ref{fp8_format} discusses the advantages of 8-bit floating point representation in handling outliers. Section .\ref{sec:quant_flow} introduces the quantization workflow and components of a standard, extended quantization scheme and a framework for tuning model performance. Section \ref{sec:results} outlines the experimental setup, presents accuracy results, and offers discussion on performance tuning. Section \ref{sec:summary} presents the conclusions and future work.

\section{Background}
\label{fp8_format}

\textbf{FP8 Value Distribution and Quantization Error: }Floating-point formats can express a large dynamic range of values using a combination of a mantissa and an exponent. A set of floating point numbers in $X \in \mathbb{R}$ are expressed as follows:  
\begin{equation}
\begin{split}
x = (-1)^s \times 2^{2^e-b} \times ( 1+ f_1\times{2^{-1}}+ f_2 \times {2^{-2}} \\
+...+ f_m\times{2^{-m}} )
\end{split}
\end{equation}

where $s\in \{0,1\}$ is the sign, \emph{e} is exponent bit width and \emph{$f_i \in \{0,1\}$} is the \emph{m}-bit mantissa or fraction.

\begin{figure*}[h!]
    \centering
    \includegraphics[scale=0.5]{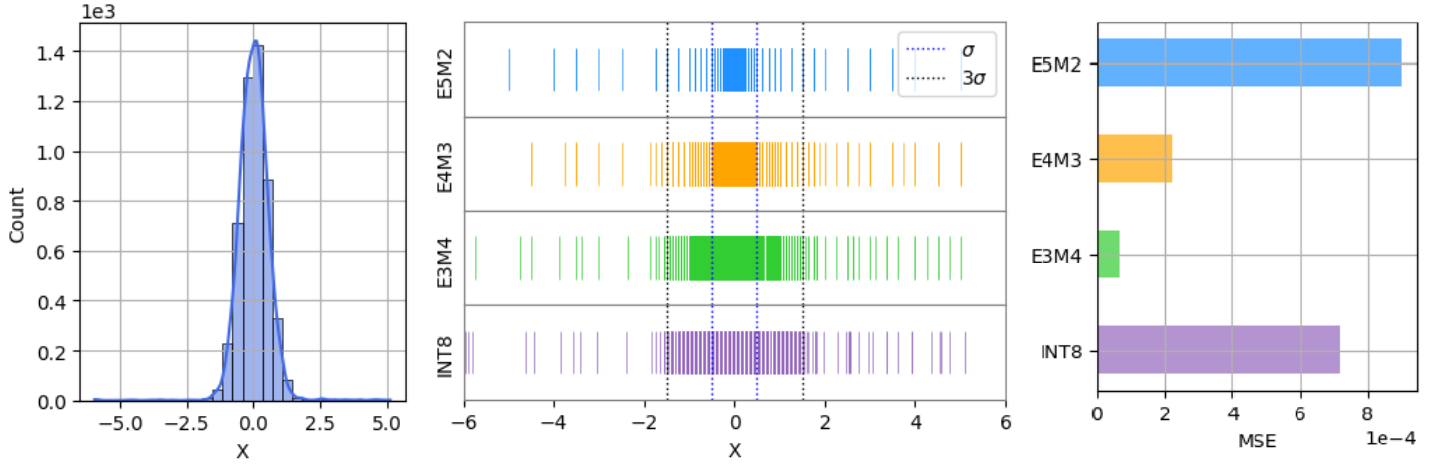}
    \caption{(\textbf{left}) Histogram of the tensor $X \sim \mathcal{N}(\mu=0.0,\,\sigma^{2}=0.5)$, that contains a small number (~1\%) of outliers uniformly distributed between -6.0 to 6.0. (\textbf{center}) Distribution of quantized  values represented by E5M2, E4M3, E3M4 and INT8 data formats. (\textbf{right}) Overall quantization error as measured by mean-square-error (MSE).  
    }
    \label{fig:fp8_format_dist}
    \vskip 0.15in
\end{figure*}

The dynamic range of a floating point format is determined by the width of its exponent. The exponent value is expressed in powers of 2 and serves as a scaling factor for the mantissa. This means that floating-point numbers are not uniformly spaced, but have a smaller step-size around zero that increases with the magnitude of the represented value. This allows floating-point formats to represent smaller values with better accuracy.

The width of the mantissa determines the number of grid points represented for each incremental step of the exponent, which in turn affects the precision of the format. These properties allow floating-point formats to support higher dynamic range without compromising the accuracy of smaller values, making them well-suited for representing many frequently occurring data patterns in deep learning workloads that exhibit long-tailed normal distributions. 
\begin{figure*}[h!]
    \centering
    \includegraphics[scale=0.7]{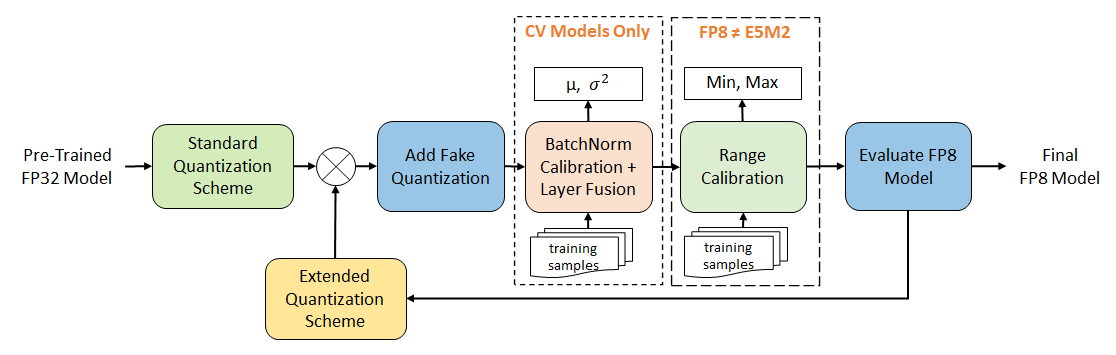}
    \caption{\emph{Standard Quantization Scheme}: default configuration for broad set of operations across different workloads, \emph{Extended Quantization Scheme}: configuration for additional operator coverage (Ex: LayerNorm, BatchNorm \& element-wise), mixed FP8 formats, dynamic quantization, \emph{BatchNorm Calibration}: recalibrate mean and variance parameters to recover accuracy lost due to quantization, \emph{Range calibration}: max scaling, outlier clipping (more discussions in Appendix A.1).}
    \label{fig:quant_flow}
    \vskip 0.15in
\end{figure*}

Figure~\ref{fig:fp8_format_dist} illustrates the differences in distribution of quantized values and impact of outliers on both FP8 and INT8 formats. In the center plot, FP8 formats show a greater concentration of grid points in the middle of the distribution, indicating a region of higher precision closer to zero. The high-precision band is wider for formats with more mantissa bits, allowing them to represent a greater percentage of the $3\sigma$ region of the original data with higher accuracy. In contrast, INT8 quantization operates with a \emph{fixed step-size} that is determined by the largest value present in the input data. This means that the outliers can significantly influence the step-size by stretching the quantization grid, resulting in fewer grid points under the $3\sigma$ region. This is reflected in the overall quantization error (MSE) shown on the right, where E4M3 and E3M4 formats have significantly outperformed INT8, while E5M2 performed worse because it has fewer mantissa bits.

\section{Quantization Workflow}
\label{sec:quant_flow}
There are several challenges in creating a generalized quantization scheme that can be applied to networks across multiple application domains and involves multiple data formats. The networks may have different requirements for dynamic range, precision and may contain operations that are sensitive to quantization. To facilitate generalization, the quantization scheme must be capable of supporting a broad set of common operations, while also having the ability to adapt to meet the unique requirements of various applications. Our framework accomplishes this by incorporating both a \emph{standard quantization scheme} that can be broadly applied, as well as an \emph{extended quantization scheme} that optimizes specific operations through an iterative tuning process. Figure~\ref{fig:quant_flow} depicts the high-level workflow for post-training FP8 quantization. The standard quantization scheme is the default configuration applied to common set of operators across different architectures, while the extended scheme is specific to an architecture and is applied incrementally in a feedback loop. 

The flow diagram in Figure~\ref{fig:quant_flow} also includes an additional \emph{BatchNorm Calibration} step applied only to computer vision models. \cite{sun_hybrid_2019} have shown that retuning BatchNorm parameters (\emph{mean} and \emph{variance}) to compensate for the variance shift caused by quantization, has significantly improved the inference accuracy. Additionally, please note that E5M2 uses \emph{direct quantization} and does not require \emph{Range Calibration} because it has sufficient dynamic range to handle outliers. For E4M3 and E3M4 formats, we found simple \emph{max} scaling to be sufficient for handling outliers. We also examined more sophisticated range-calibration methods such as KL divergence \cite{darvish2020pushing, 8-bit-tensorrt}, MSE error \cite{choukroun2019low, zhao2019improving} and percentile \cite{gholami2021survey} which did not provide any additional benefits.

\subsection{Standard Quantization Scheme}
\label{sec:standard_scheme}
This section outlines the components of the standard quantization scheme, which is derived from our extensive studies conducted on several deep learning tasks across multiple application domains. This scheme is applied to the common subset of operators including Convolution, Linear and Embedding. This scheme is also identical to INT8 quantization scheme, allowing a fair accuracy comparison.

\textbf{Weight and Activation Scaling:} We recommend using per-channel scaling for weights across all networks. Although FP8 formats have sufficient dynamic range to handle common weight distributions, empirical evidence suggests that applying per-channel scaling can reduce rounding errors by effectively utilizing the full encoding space for each channel. Similarly, we found per-tensor scaling to be adequate for handling outliers using FP8 formats. The scale factors are computed as below:
\begin{equation}
s =  \left( float\_max/max\_T \right)
\end{equation}
where \emph{float\_max} is the max representable value of the selected FP8 format, and \emph{max\_T} is the calibrated \emph{absmax} value of the tensor. Some recent studies \cite{xiao2022smoothquant, outlier_supression, dettmers2022llm} have indicated that per-channel activation scaling can benefit INT8 quantization. However, such methods may require special kernel implementations that are likely to incur higher compute overheads, hence they are not included in our study. 
 
\textbf{First and Last Operator:} Previous studies \cite{han2015learning, PACT, micikevicius2022fp8} on convolution networks have shown that the first convolution and the last fully-connected layers are more sensitive to quantization. These two operators typically constitute $<$ 1\% of the total computation. Therefore, we continue to maintain these layers in higher precision to preserve model accuracy. Please note that this exception is only applicable to convolutional neural networks.   

\subsection{Extended Quantization Scheme}
\label{sec:extended_scheme}
This section outlines the quantization scheme that is selectively applied to address the specific needs of an application. These methods are applied incrementally to maximize model efficiency while preserving accuracy.

\textbf{Expanded Operator Coverage:} Neural networks spend significant fraction of their execution time in memory-bound operations such as LayerNorm, BatchNorm\footnote{Ones that cannot be folded into Convolution layers, Ex: Densenet} and element-wise operators such as Add and Mul. Previous attempts \citet{layernorm_accuracy_issue, Ibert} to quantize these operators using integer approximation were unsuccessful in maintaining the model accuracy. Our experiments show that FP8 formats are capable of handling these operators without sacrificing model accuracy.

\begin{figure*}[htbp]
    \centering
    \includegraphics[scale=0.38]{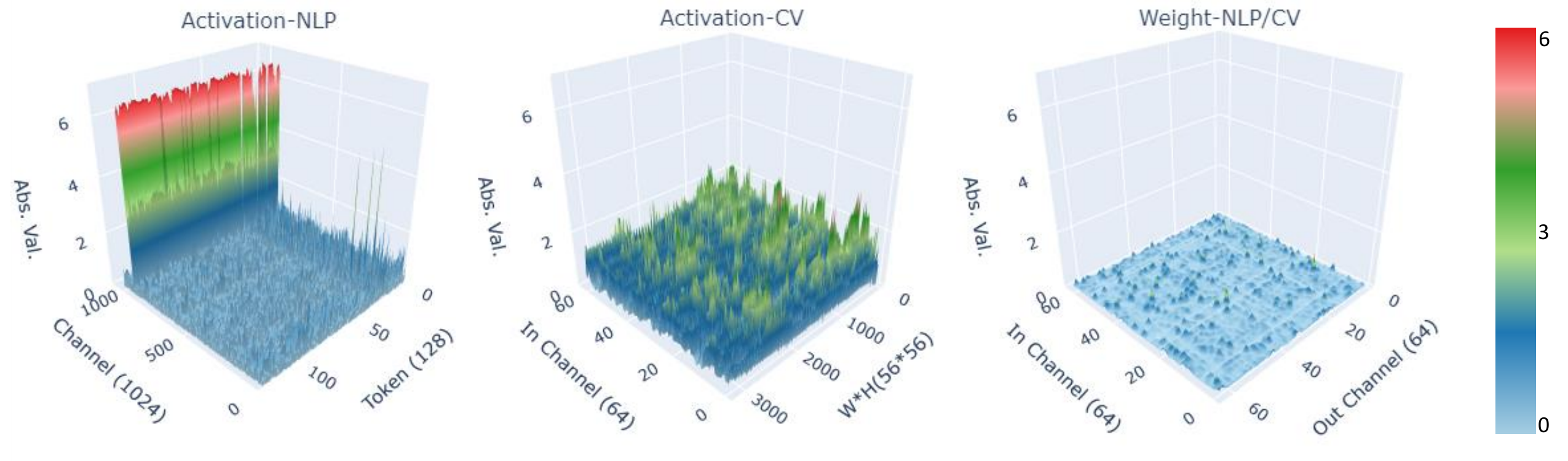}
    \caption{Tensor Distributions: (\textbf{left}) activations in NLP workloads contain outliers, hence they are \emph{range-bounded}, (\textbf{center}) Activation in CV workloads tend to be \emph{precision-bounded}, (\textbf{right}) Weight tensors from  both CV \& NLP networks tend to be \emph{precision-bounded}.}
    \label{fig:hfp8_tensor}
    \vskip 0.15in
\end{figure*}

\textbf{Mixed FP8 Formats:} The data distributions of weights and activations can vary depending on the architecture of the model and the dataset it is trained on. Figure~\ref{fig:hfp8_tensor} shows typical distributions of weight and activation tensors in NLP and computer vision workloads. The weight distributions in both classes of models tend to follow normal distributions with lots values near zero. 
These tensors require more mantissa bits in the data format to represent the distribution accurately. In contrast, activations of NLP models show a lot of outliers which demand a larger dynamic range in the data format to ensure the outliers are accurately represented. We balance this trade-off by assigning E5M2 or E4M3 format for \emph{range-bound} tensors and E3M4 for \emph{precision-bound} tensors. 

\textbf{Static vs. Dynamic Quantization:} We use static quantization as the default method throughout our study because it is computationally more efficient. However, we studied the accuracy impact of dynamic quantization on all FP8 formats and found that it offers no additional benefits to E5M2 but observed a noticeable improvement in accuracy for E4M3 and E3M4 formats on selected models.

\section{Results}
\label{sec:results}

\subsection{Experimental Setup}
\label{sec:exp_setup}

We demonstrate the FP8 quantization results using a software emulation framework which contains two major components, \emph{data type emulation} and \emph{model quantization}. For data type emulation, we utilized the \href{https://github.com/IntelLabs/FP8-Emulation-Toolkit}{FP8 Emulation Toolkit}, which provides a reference implementation that runs FP32 hardware. We leverage \href{https://github.com/intel/neural-compressor}{Neural Compressor} to perform model quantization by incorporating both standard and extended quantization schemes, along with FP8 specific quantization methods such as BatchNorm calibration and support for mixed FP8 formats. Our framework supports a wide range of quantized operators, including compute operators such as Convolution, Linear, MatMul, BatchMatMul and memory operators such as Embedding, BatchNorm, LayerNorm, Add and Mul.

\begin{table*}[h!]
    \centering
    \caption{Workload Pass Rate. The \textbf{bold} shows the overall highest pass rate where E4M3 is 92.64\% and INT8 is 65.87\%. In particular, E4M3 shows the promising workload coverage 96.32\% on NLP.}
    \vskip 0.15in
    \begin{tabular}{lcccc}
    \toprule
        Data Type & Quantization Approach & Pass Rate (CV) & Pass Rate (NLP) & Pass Rate (All) \\
    \midrule
        E5M2 & Direct & 55.26\% & 78.42\% & 74.89\% \\
        E4M3 & Static & 73.68\% & \textbf{96.32}\% & \textbf{92.64}\% \\ 
        E4M3 & Dynamic & 71.05\% & 92.11\% & 88.74\% \\
        E3M4 & Static & \textbf{78.95}\% & 92.11\% & 90.04\% \\
        E3M4 & Dynamic & \textbf{78.95}\% & 92.11\% & 90.04\% \\
        INT8 & Static CV $|$ Dynamic NLP & 57.89\% & 67.65\% & 65.87\% \\ 
    \bottomrule
    \end{tabular}
    \label{tbl:pass_rate}
    \vspace{0.15in}
\end{table*}

\begin{figure*}[htbp]
    \centering
    \includegraphics[scale=0.6]{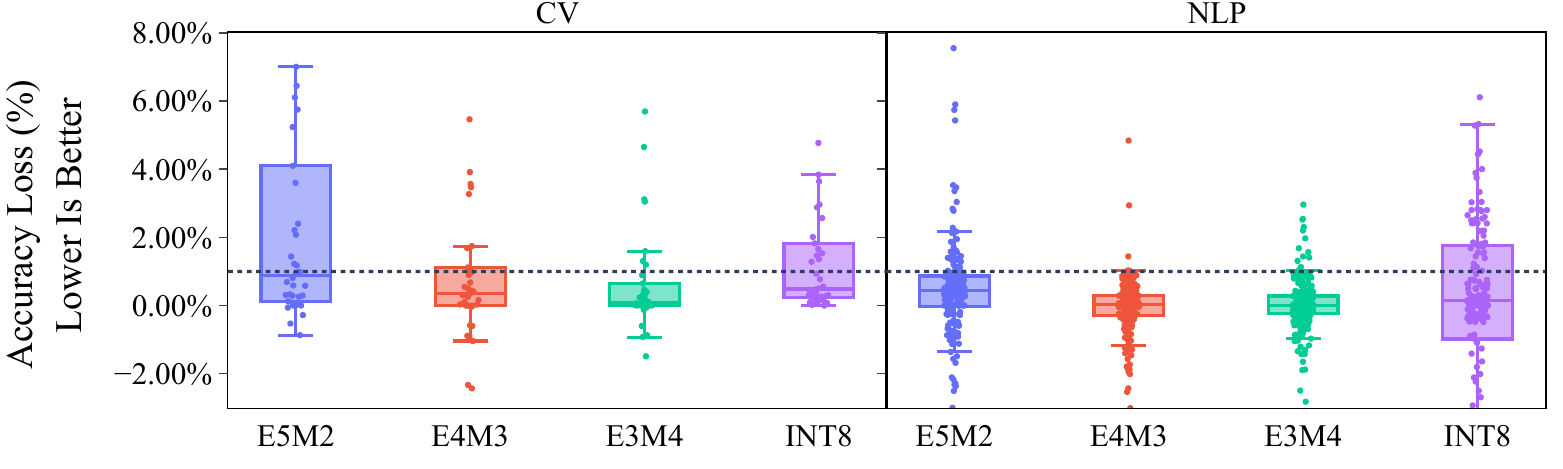}
    \caption{Variability in accuracy loss: INT8 shows higher variability for CV models than E4M3 and E3M4 due to its ineffectiveness on models such as EfficientNet, MobileNetV3, and ViT. Quantization-aware training may partially mitigate this issue, but it is out of scope of this paper. E4M3 and E3M4 show better accuracy \& less variability with very few outliers compared to INT8.}
    \label{fig:fp8_results}
    \vskip 0.1in
\end{figure*}
\begin{table*}[h!]
    \caption{Model Accuracy. The \textbf{bold} shows the best accuracy is less than 1\% loss against FP32 baseline.}
\centering
\setlength{\tabcolsep}{3.2mm}
\vskip 0.15in
\begin{tabular}{lcccccc}
\toprule
Model & Dataset/Task & FP32 & E5M2 & E4M3 & E3M4 & INT8 \\
\midrule
ResNet-50       & ImageNet 2012   & 0.7615                     & 0.7544            & 0.7592                     & \textbf{0.7604}            & 0.7595            \\
DenseNet-121    & ImageNet 2012   & 0.7444                     & 0.7435            & 0.7451                     & \textbf{0.7459}            & 0.7253                     \\
Wav2Vec2        & LibriSpeech     & 0.9660                     & 0.9632            & \textbf{0.9661}                     & 0.9658            & 0.9552                     \\
DLRM            & Criteo Terabyte & 0.8027                     & 0.8016            & \textbf{0.8025}                     & \textbf{0.8025}            & 0.8024            \\
Bert-Base       & STS-B           & 0.8975                     & 0.8934            & \textbf{0.8979}                     & 0.8966            & 0.8809                     \\
Bert-Large      & COLA            & 0.6257                     & 0.6238            & 0.6257                     & 0.6282            & \textbf{0.6389}                     \\
DistilBert      & MRPC            & 0.8916                     & 0.8897            & 0.8943                     & 0.895             & \textbf{0.9042}                   \\
Bloom-7B1 & Lambada-openai  & 0.5764 & 0.5424 & 0.5748 & 0.5824 & \textbf{0.5977} \\
Bloom-176B & Lambada-openai & 0.6777 & 0.6753 & 0.6757 & \textbf{0.6938} & 0.6899 \\
LLaMA-65B & Lambada-openai & 0.7908 & 0.7840 & \textbf{0.7914} & 0.7778 & 0.7155 \\
\bottomrule
\end{tabular}
\vskip 0.15in
\label{tbl:quant_results}
\end{table*}
We evaluated our quantization methods on more than 200 different tasks, using 75 unique model architectures and over 20 different datasets. The models were selected randomly from a pool of a combination of diversity and popularity from mainstream hubs such as \href{https://huggingface.co/models}{Hugging Face Models} and \href{https://github.com/pytorch/vision}{Torch Vision}, as well as individual models from Github based on their popularity. The following is a partial list of workloads that are broadly categorized under Natural Language Processing (NLP) and Computer Vision (CV).

\textbf{Text and Natural Language Processing}: We have evaluated 38 different networks in this category on a wide range of NLP tasks, which can be further subdivided as follows:
\vspace{-5pt}
\begin{itemize}
    \setlength{\itemindent}{0em}
    \setlength{\itemsep}{0em}
    \item \emph{Generative language modeling}. We evaluated \emph{Bloom}~\cite{scao2022bloom} and \emph{LLaMA}~\cite{touvron2023llama}, two representative open-source LLMs, and evaluate the accuracy using \href{https://github.com/EleutherAI/lm-evaluation-harness}{\emph{lambada-openai}}. 
    \item \emph{Text classification}. We evaluated over 30 different networks (e.g, \emph{Bert-Large}~\cite{devlin2018bert}, \emph{DistilBert}~\cite{sanh2019distilbert}, \emph{Longformer}~\cite{beltagy2020longformer}) on a wide variety of tasks (e.g., \emph{mrpc}, \emph{cola}, \emph{sts-b}, \emph{sst2}).
    \item \emph{Summarization}. We measured the accuracy of \emph{pegasus}~\cite{zhang2020pegasus} on \emph{samsum} dataset.
    \item \emph{Other NLP tasks}. Few other selected models such as MarianMT~\cite{mariannmt} for neural machine translation and DialogGPT~\cite{zhang2019dialogpt} for language modeling on WMT\_EN\_RO and \href{https://huggingface.co/datasets}{wikitext} datasets.
\end{itemize}

\vspace{-5pt}
\textbf{Image and Computer Vision}: We evaluated 34 different networks on various computer vision tasks from the following categories.
\vspace{-4pt}
\begin{itemize}
    \setlength{\itemindent}{0em}
    \setlength{\itemsep}{0em}
\item \emph{Image generation}. We evaluated Stable Diffusion, an open-source state-of-the-art latent text-to-image diffusion model and evaluate using FID~\cite{fid}. 
\item \emph{Image classification}. We evaluate a wide range of convolutional neural networks (CNNs) such as VGG~\cite{simonyan2014very}, GoogleNet~\cite{szegedy2015going}, ResNet~\cite{he2016deep}, ShuffleNet~\cite{zhang2018shufflenet}, EfficientNet~\cite{tan2019efficientnet}, and Transformer-based vision models such as ViT~\cite{dosovitskiy2020image} on \href{https://www.image-net.org/challenges/LSVRC}{ImageNet ILSVRC 2012} and \href{https://www.cs.toronto.edu/~kriz/cifar.html}{CIFAR-10}.
\item \emph{Image segmentation \& object detection}. We select typical models such as U-Net~\cite{ronneberger2015u} for image segmentation using the dataset from Kaggle Carvana Image Masking Challenge~\cite{kaggle-dataset} and YoloV3~\cite{redmon2018yolov3} for object detection using COCO2014~\cite{coco2014}.
\end{itemize}
\vspace{-4pt}
\textbf{Audio and Speech Processing}. We evaluated two models HuBERT~\cite{hsu2021hubert} and wav2vec 2.0~\cite{baevski2020wav2vec} for speech recognition and evaluate the accuracy using LibriSpeech~\cite{librispeech}.

\textbf{Recommendation System}. We evaluated Deep Learning Recommendation Model (DLRM)~\cite{naumov2019deep} and measured the accuracy on \href{https://ailab.criteo.com/ressources/criteo-1tb-click-logs-dataset-for-mlperf}{Criteo Terabyte}.


\subsection{Quantization Results}
\label{sec:quant_results}

\subsubsection{Accuracy} 
\label{sec:accuracy}


\begin{figure*}[h!]
    \centering
    \includegraphics[scale=0.32]{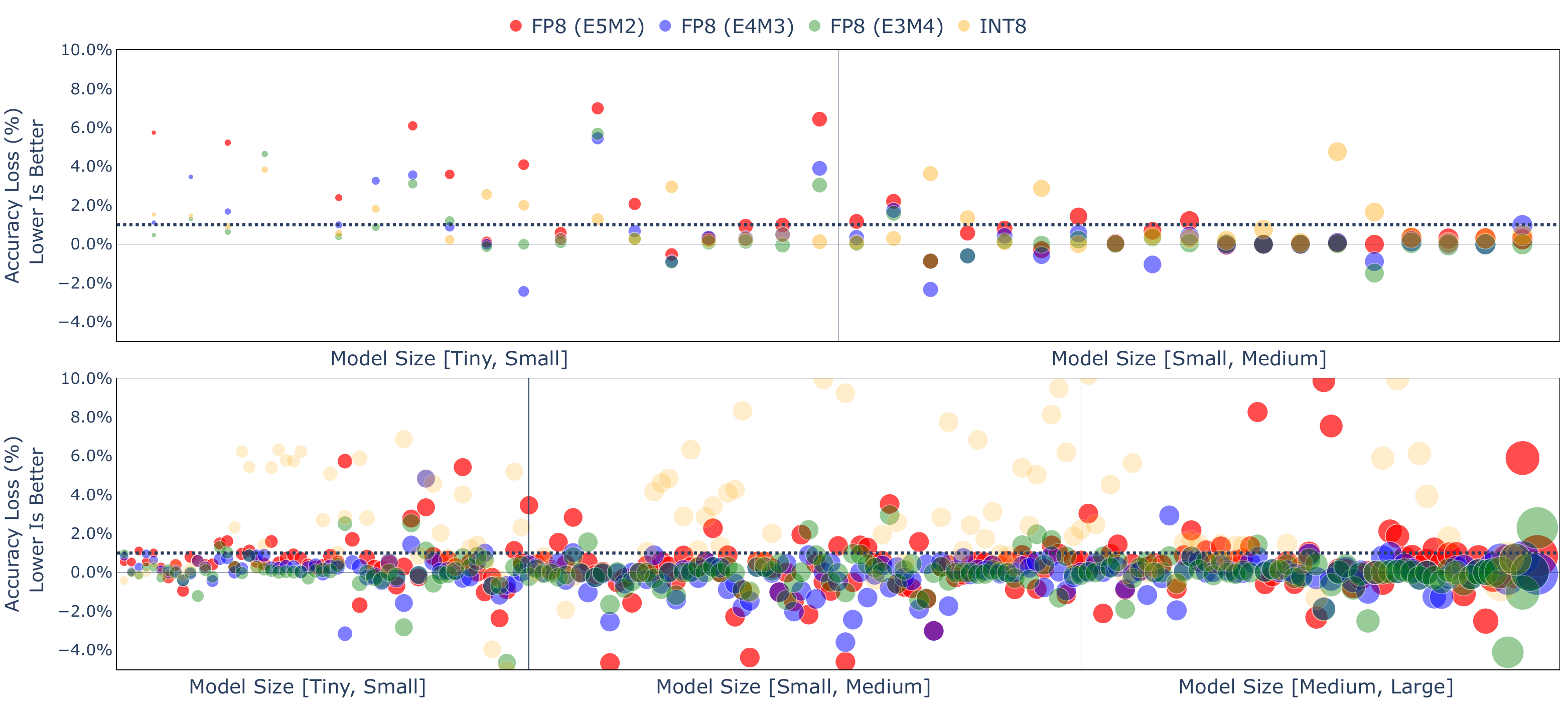}
    \caption{Accuracy Loss by Size on CV (top) and NLP (bottom). The model size is represented by the ball size in the scale of $log10(model\_size)$, where tiny/small/medium/large is defined by the size range in MB $<=32$, $(32, 384]$, $(384, 512]$, and $> 512$ respectively. Note that some points are overlayed due to the similar accuracy (e.g., E4M3 in blue and E3M4 in green on NLP models).}
    \label{fig:accuracy_ball_size}
    \vskip 0.15in
\end{figure*}

Note that the \emph{pass rate} in Table~\ref{tbl:pass_rate} is the percentage of workloads that meet the accuracy criterion of 1\% relative loss against FP32 baseline. SmoothQuant~\citet{xiao2022smoothquant} is enabled on NLP models with the default smoothing alpha value (alpha tuning is out of scope in this paper). Figure~\ref{fig:fp8_results} illustrates the variability of accuracy loss for different data formats across CV and NLP workloads. 

Table~\ref{tbl:quant_results} shows the accuracy of a few representative samples from all CV and NLP workloads. Figure~\ref{fig:accuracy_ball_size} shows the accuracy loss of all workloads sorted by the model size in ascending order. 

\begin{figure*}[h!]
    \centering
    \includegraphics[scale=0.43]{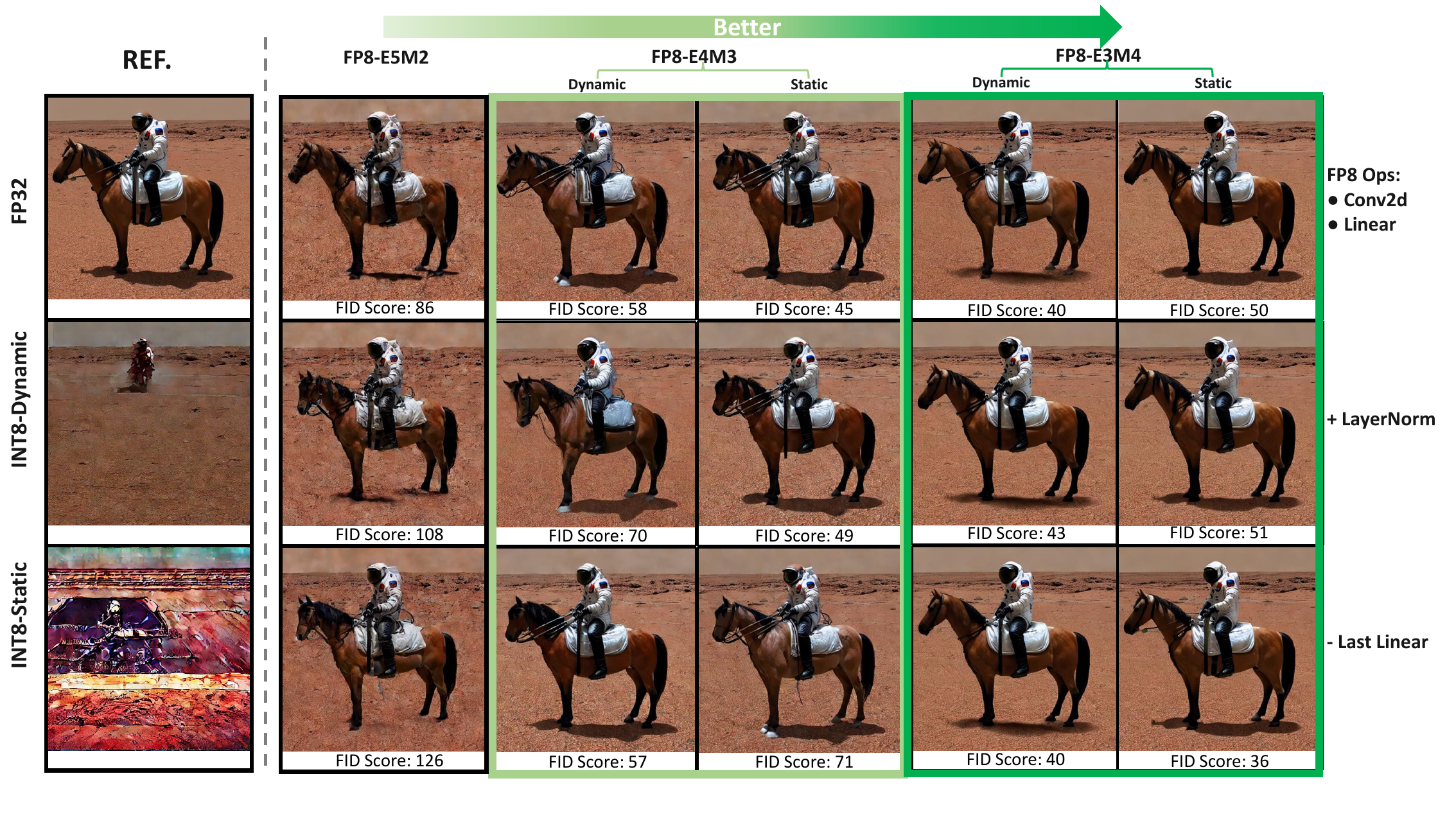}
    \vspace{-4mm}
    \caption{Stable Diffusion with Prompt "A photo of an astronaut riding a horse on Mars"}
    \label{fig:hfp8_stable_horse}
    \vskip 0.15in
\end{figure*}

\begin{figure*}[h!]
    \centering
    \includegraphics[scale=0.55]{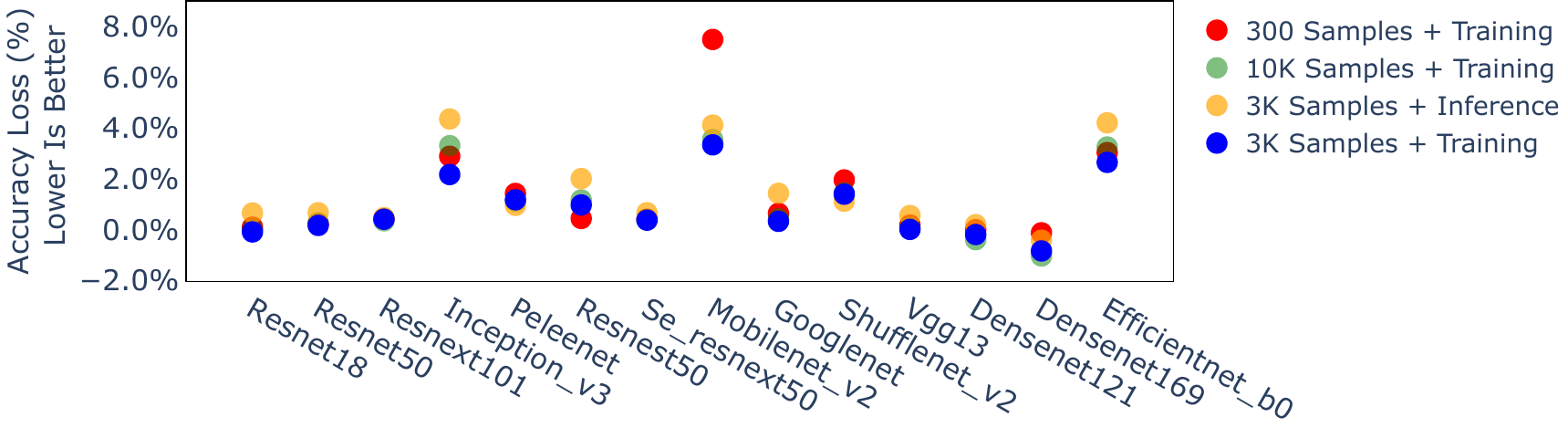}
    \caption{CV Models with BatchNorm Operation}
    \label{fig:fp8_batchnorm}
    \vskip 0.15in
\end{figure*}

\begin{table*}[h!]
	\centering
	\caption{Generated Text of Bloom on 32 Input Tokens}
        \vskip 0.15in
	\begin{tabularx}{\textwidth}{XXX}
		\toprule
		  \textbf{Prompt}: Once upon a time, there existed a little girl, who liked to have adventures. She wanted to go to places and meet new people, and have fun. \\
            \midrule
            \textbf{Output (FP32)}: One day, she decided to go on a trip. She packed her suitcase and went to the airport. When she got there, she found out that there was no flight to her destination, so she decided to take a bus. When she got there, she found out that there was no bus to her destination...\\
            \midrule
    	\textbf{Output (INT8)}: This little girl was very adventurous. One day she decided to go on a trip to a faraway country. When she got there the little girl \emph{saw many strange} things. She \emph{saw many strange} people. She \emph{saw many strange} animals. She saw many strange sights. She saw many ...\\
		\midrule
            \textbf{Output (E3M4)}: One day, she decided to go on an adventure. She packed her suitcase and went to the airport. She boarded a plane and flew to New York City. There, she met a man, and they had a great time together. They went to a restaurant and ate delicious food. Then, they went to... \\
		\bottomrule
	\end{tabularx}
 \label{tab:text_generation}
 \vskip 0.15in
\end{table*}


\subsubsection{Generation Quality} 
Figure~\ref{fig:hfp8_stable_horse} shows the image generated by Stable Diffusion with the prompt "A photo of an astronaut riding a horse on Mars". Our subjective analysis reveals that FP8 formats achieve superior image quality compared to INT8, as indicated by the green arrow. Additionally, E4M3 and E3M4 produce smoother images and generate more intricate details, particularly on the astronaut. We employ FID score to compare the quality of generated images (lower is better) and see that FID score aligns with our subjective evaluation. More samples on Stable Diffusion are shown in Appendix A.2.

Table~\ref{tab:text_generation} shows the sample text generated by Bloom on the prompt with 32 input tokens using beam search size 4. Given the prompt as the input, you can see E3M4 shows better response than INT8 with more comprehensive content and few repeated tokens (e.g., \emph{saw many strange}). Appendix A.3 shows the full output on different data format and quantization approach.

\subsection{Discussion}
\label{sec:discussions}

\subsubsection{Standard Quantization Scheme}
\textbf{Quantizing First and Last Operators :}
For convolutional networks, quantizing the first and last operators reduced the \emph{Pass Rate} for E5M2 and E4M3 formats by 25\% and 15\% respectively. However, E3M4 can maintain a \emph{Pass Rate} of 70\% even with the first and last operators quantized. Therefore, we recommend the enabling of first and last operators for FP8 quantization as a tuning option.

\textbf{BatchNorm Calibration:} We use data augmentation to enhance the feature diversity of the calibration data which impacts the quality of BatchNorm statistics and model accuracy. Figure~\ref{fig:fp8_batchnorm} compares the effectiveness of training and inference data augmentation methods in preserving model accuracy at different calibration data sample sizes. We found training transform to be more effective even at smaller sample sizes ($<$3K). However, we recommend sample size of 3K with training transform for achieving best results across a wide range of networks. 

\vspace{-3pt}
\begin{figure*}[h!]
    \centering
    \includegraphics[scale=0.6]{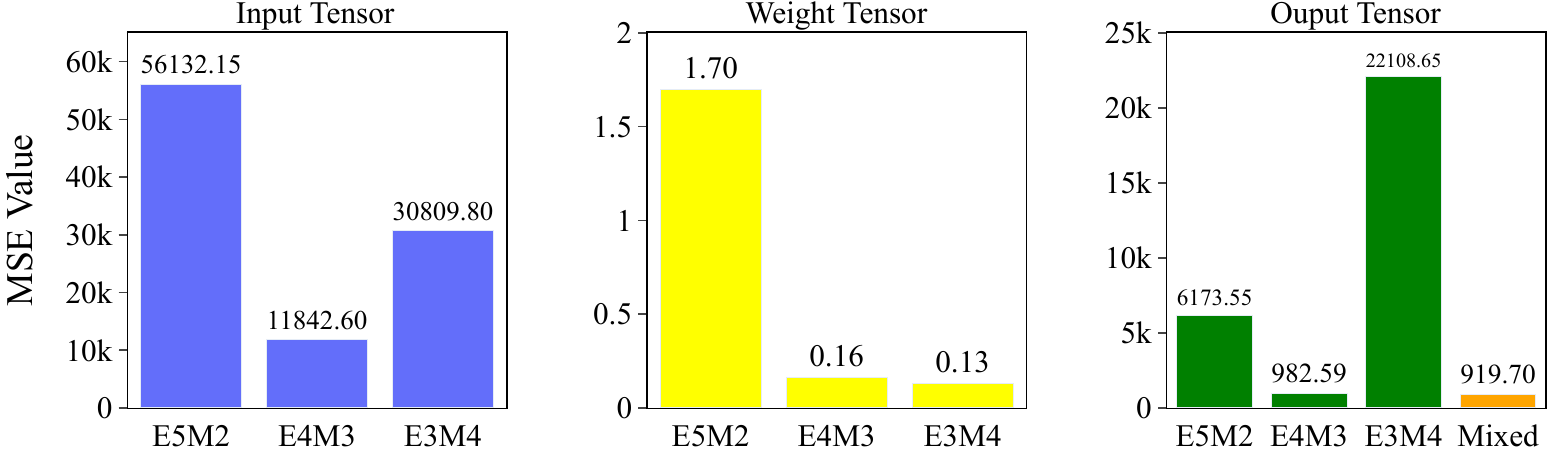}
    \caption{MSE of FP8 Quantization with Mixed Formats vs. Single Format on Bert-Base (MRPC)}
    \label{fig:mixed_fp8_type}
\end{figure*}
\begin{table*}[h!]
    \caption{Model Accuracy of FP8 Format (Single vs. Mixed). Mixed FP8 formats (in bold) show higher accuracy than all the other single FP8 formats on the below NLP workloads.}
\centering
\setlength{\tabcolsep}{4.5mm}
\vskip 0.2in
\begin{tabular}{lcccccc}
\toprule
Model & Task & FP32 & E5M2 & E4M3 & E3M4 & Mixed \\
\midrule
Bert-Base & MRPC & 0.9069 & 0.9040 & 0.9050 & 0.9050 & \textbf{0.9069} \\
Bert-Large & RTE & 0.7256 & 0.6968 & 0.7329 & 0.6931 & \textbf{0.7365} \\
Funnel & MRPC & 0.9225 & 0.9215 & 0.9207 & 0.3704 & \textbf{0.9233} \\
Longformer & MRPC & 0.9146 & 0.8374 & 0.9113 & 0.9084 & \textbf{0.9143} \\
\bottomrule
\end{tabular}
\label{tbl:mixed_fp8_formats}
\end{table*}
\begin{table*}[h!]
    \caption{Model Accuracy of Quantization Approach (Static vs. Dynamic)}
\centering
\setlength{\tabcolsep}{3.1mm}
\vskip 0.2in
\begin{tabular}{lccccc}
\toprule
Model & Task & FP8 Format & Dynamic & Static & Improvement \\
\midrule
Bert-Base & MRPC & E4M3 & 0.9151 & 0.9072 & \textbf{+0.87\%} \\
Bert-Base & COLA & E4M3 & 0.6058 & 0.6033 & \textbf{+0.41\%} \\
Bert-Large & RTE & E4M3 & 0.7401 & 0.7329 & \textbf{+0.98\%} \\
Xlm-Roberta-Base & MRPC & E3M4 & 0.8962 & 0.8919 & \textbf{+0.48\%} \\
\bottomrule
\end{tabular}
\label{tbl:static_dynamic}
\end{table*}
\subsubsection{Extended Quantization Scheme}
\label{sec:extended_quantization} 
\textbf{Mixed FP8 Formats:} Figure~\ref{fig:mixed_fp8_type} illustrates how using mixed FP8 formats on the input can impact the quantization error of the output of a Linear operator from BERT-base (MPRC) model. Our experiments show that using E4M3 for activations and E3M4 for weights produced best accuracy results on a range of NLP workloads. The accuracy improvements achieved by this scheme for Bert, Funnel, and Longformer models are presented in Table~\ref{tbl:mixed_fp8_formats}.

\begin{figure*}[h!]
    \vskip 0.25in
	\centering
	\subfigure[CV Models]{
		\begin{minipage}{\textwidth} 
                \centering
               \includegraphics[scale=0.429]{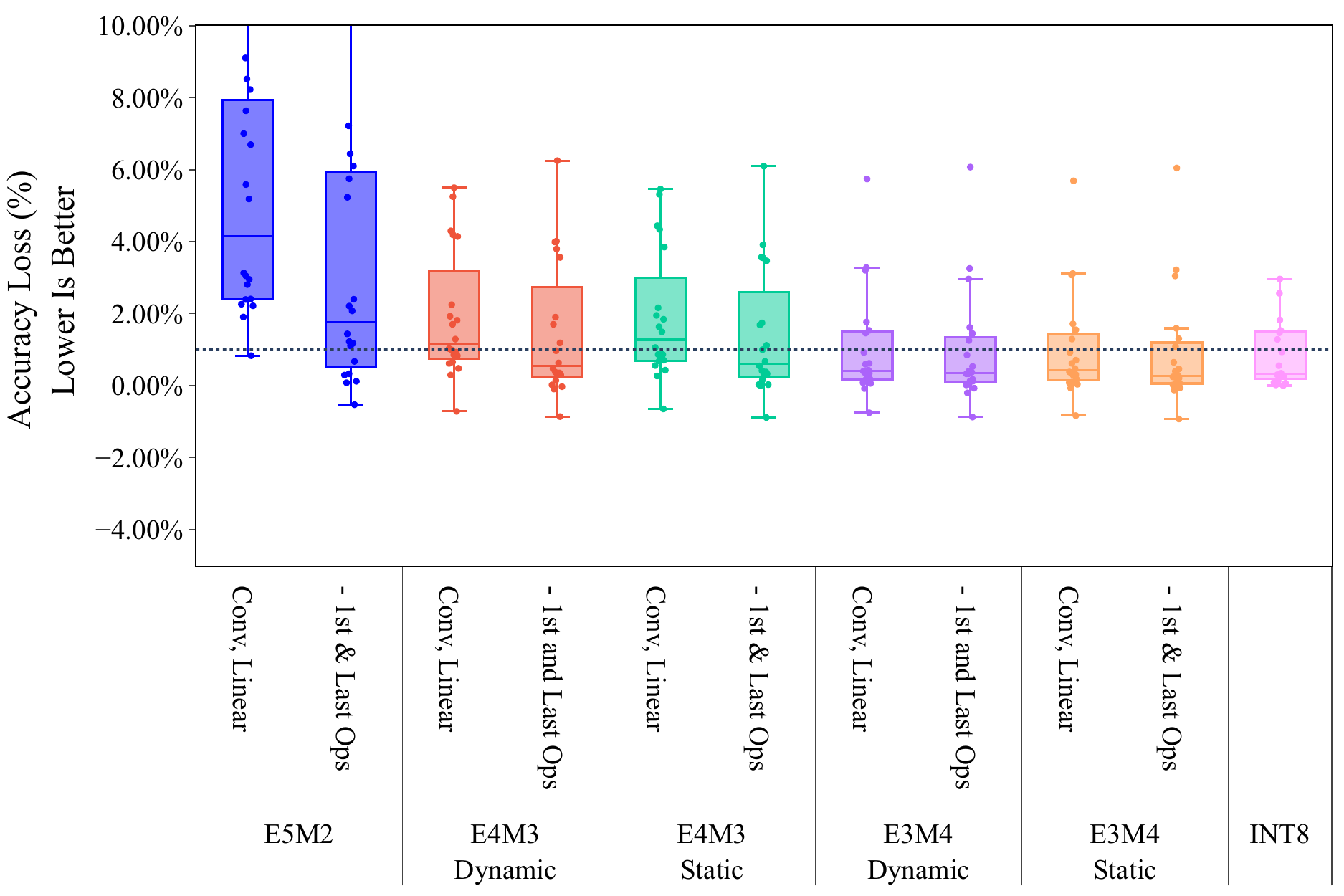} \\
		\end{minipage}
        }
	\subfigure[NLP Models]{
		\begin{minipage}{\textwidth}
                \centering
                \includegraphics[scale=0.429]{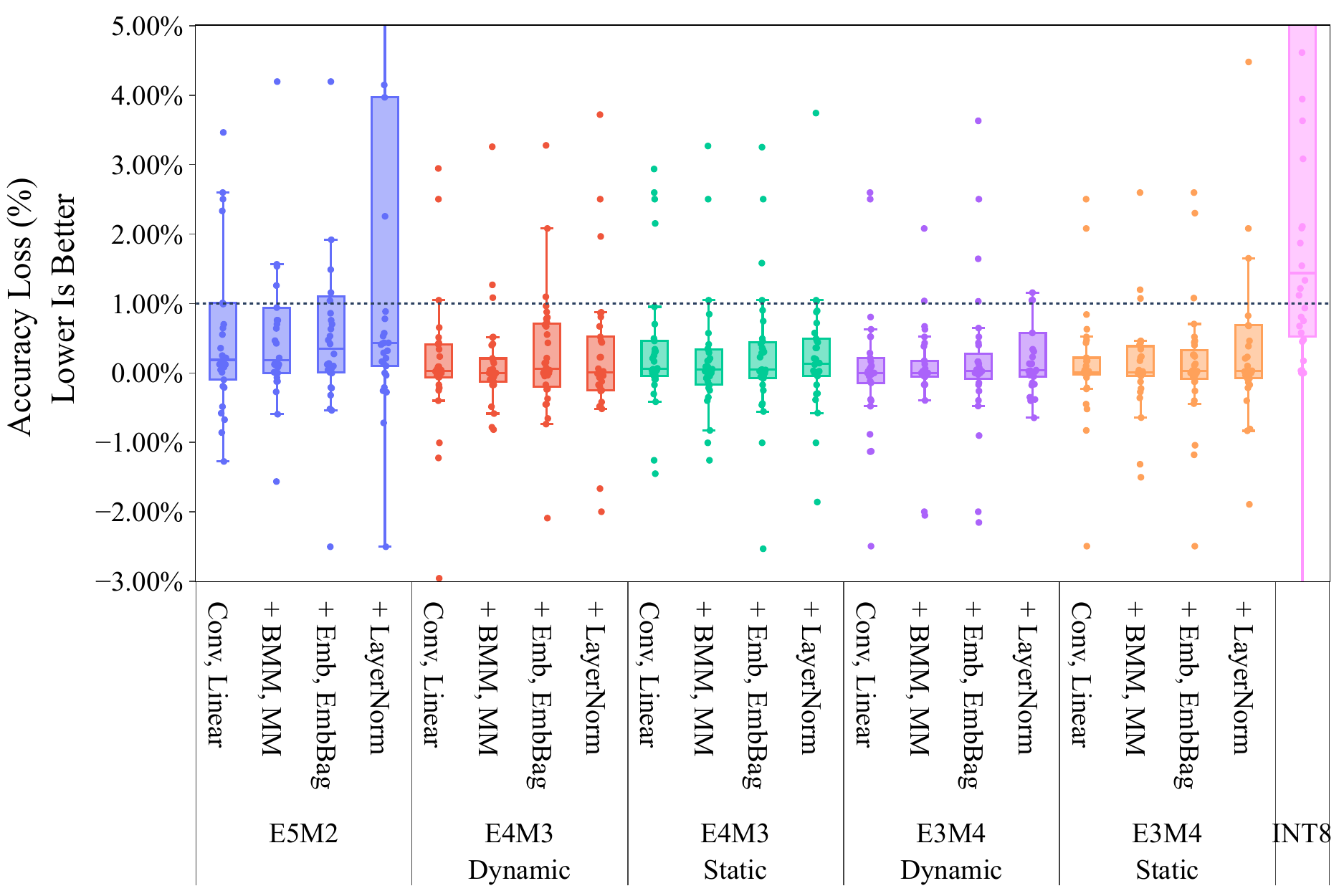} \\
		\end{minipage}
	}
 	\caption{Model Accuracy Impact by Extended Quantization Recipes} 
	\label{fig:extended_quant_recipes}
    \vskip 0.35in
\end{figure*}

\textbf{Expanded Operator Coverage:} Figure~\ref{fig:extended_quant_recipes} has the results from our quantization studies extended to a wider range of operators such as BatchMatMul, MatMul, Embedding and LayerNorm. Our results show that E4M3 achieves overall better accuracy and smaller variability in accuracy loss across a broad range of NLP tasks.
\textbf{Static vs. Dynamic Quantization:} While static quantization is the default approach in our recipes, we also studied the impact of dynamic quantization on model accuracy. The results indicate that dynamic quantization can improve the accuracy of NLP models when quantizing with E4M3 and E3M4 formats as shown in Table~\ref{tbl:static_dynamic}.

\section{Summary and Future Work}
\label{sec:summary}
We present a set of post-training quantization recipes for FP8 inference and demonstrate the effectiveness across 75 unique network architectures covering a wide range of tasks such as language modeling, text generation, image classification and generation. We recommend E3M4 and E4M3 as the default FP8 format for CV and NLP models respectively, while additional recipes such as mixed FP8 formats and expanded FP8 operator coverage are worthwhile exploring to produce an optimal FP8 model. As our future work, we plan to apply FP8 quantization recipes to more diverse LLM models (e.g., BioGPT~\cite{luo2022biogpt}, Llama2 Chat~\cite{touvron2023llama}, Code Llama~\cite{roziere2023code}), and contribute our recipes and implementation to the open source community.

\nocite{langley00}

\bibliography{paper}
\bibliographystyle{mlsys2024}

\appendix
\section{Appendix}

\subsection{Range Calibration Algorithms}
\label{sec:appendix_cal_algos}

Scale algorithm is not applied for E5M2 due to its large dynamic range. However, the scale is very crucial for E4M3 to help make sure all data is recorded in E4M3 data range. Mainly, there are three classic scale algorithms used in INT8 quantization. Percentile and KL can help INT8 clip the min-max of observed data range to skip outliers and improve the accuracy of data representation\cite{jiang2021efficient}. However, they may have different behavior on FP8 due to the special data distribution of FP8.

\begin{figure}[htbp]
    \centering
    \includegraphics[scale=0.4]{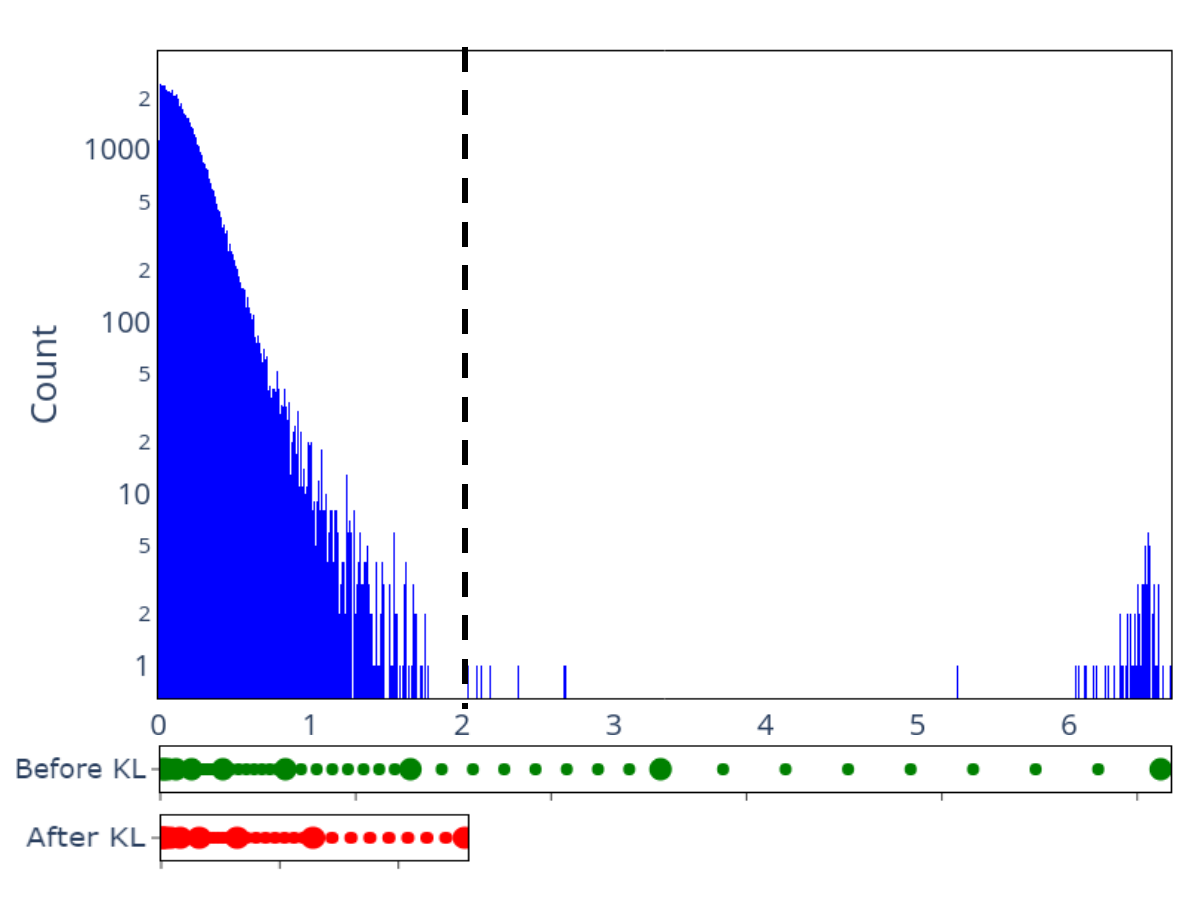}
    \caption{A KL Demo for FP8 mapping}
    \label{fig:kl_demo}
\end{figure}

In Figure~\ref{fig:kl_demo}, we show a demo to explain the shortage of KL when using FP8. The demo uses a tensor with some outliers around 6 and after KL process, the clipped max value is 2. The lines at the bottom show FP8 mapped data with different max values. The upper line have a large data range from 0-6 while the other line have more representations for small values. We expect the lower line have a better representation than the upper one, but it actually have a large MSE than the upper one. We can observer that the density of the last block in the lower line is much sparse than the upper one, while the enhanced small value representations do not help a lot in MSE.

As mentioned early, the FP8 has advantages of representing larger range of values and obtaining better accuracy at lower range because of denser representation on the contrary to the uniform representation at the whole range of INT8. FP8 format is represented by exponent bits (e) and mantissa bits (m). 
Here, we use E(e)M(m) as FP8 representation to demonstrate our point. To calculate the density of number for E(e)M(m), we choose a simplified method that uses the differentials between two points with exact same mantissa of value 1 but with a difference of 1 in exponent as $[1\times2^n, 1\times2^{n+1})$. We know that for any range with such endpoints, the number of values being represented is always $2^m$. Therefore, we can calculate the density on this range is as: 
\begin{equation}
    D_{E(e)M(m)}=2^m/(2^{n+1}-2^n )=2^{m-n}
\end{equation}

As is well known, any decimal number N can be represented by binary number with exponent $Floor[log_2N]$. Hence, the density of E(e)M(m) representation in decimal system is: 

\begin{equation}
D_{E(e)M(m)}=2^{m-Floor[log_2N]}   
\end{equation}

It’s clearly shown that the smaller the number N the denser the number of values being represented. On the contrary, the larger the number N the sparser the number of values being represented. Therefore, we always prefer to examine the histogram of our tensor’s value and make sure always to represent the high frequency part of our tensor on the lower range on FP8 with higher density, which is in sharp contrast to INT8 with uniform density. Also shown in the density expression, the more the mantissa the denser the number of values being represented as expected.  

Operator level means we have to fallback some operators to high precision to let the quantized model meed the accuracy goal. Theoretically, the more operators converted to low precision, the worse the precision will be. Usually, there are special operator types that have a big impact on accuracy, such as LayerNorm. Also, there are some individual operators that have the most impact on accuracy, such as the first and last operators.

The tuning strategy we proposed allows an automatic tuning for the best accuracy, performance or Pareto optimal. The search space is based on the combination of all tune-able parameters by default. Typically, a customized search space based on our experiment result can help narrow down the search space.

\subsection{More Image Generation Samples from Stable Diffusion}
\label{sec:more_samples}

\begin{figure*}[h!]
    \centering
    \includegraphics[scale=0.45]{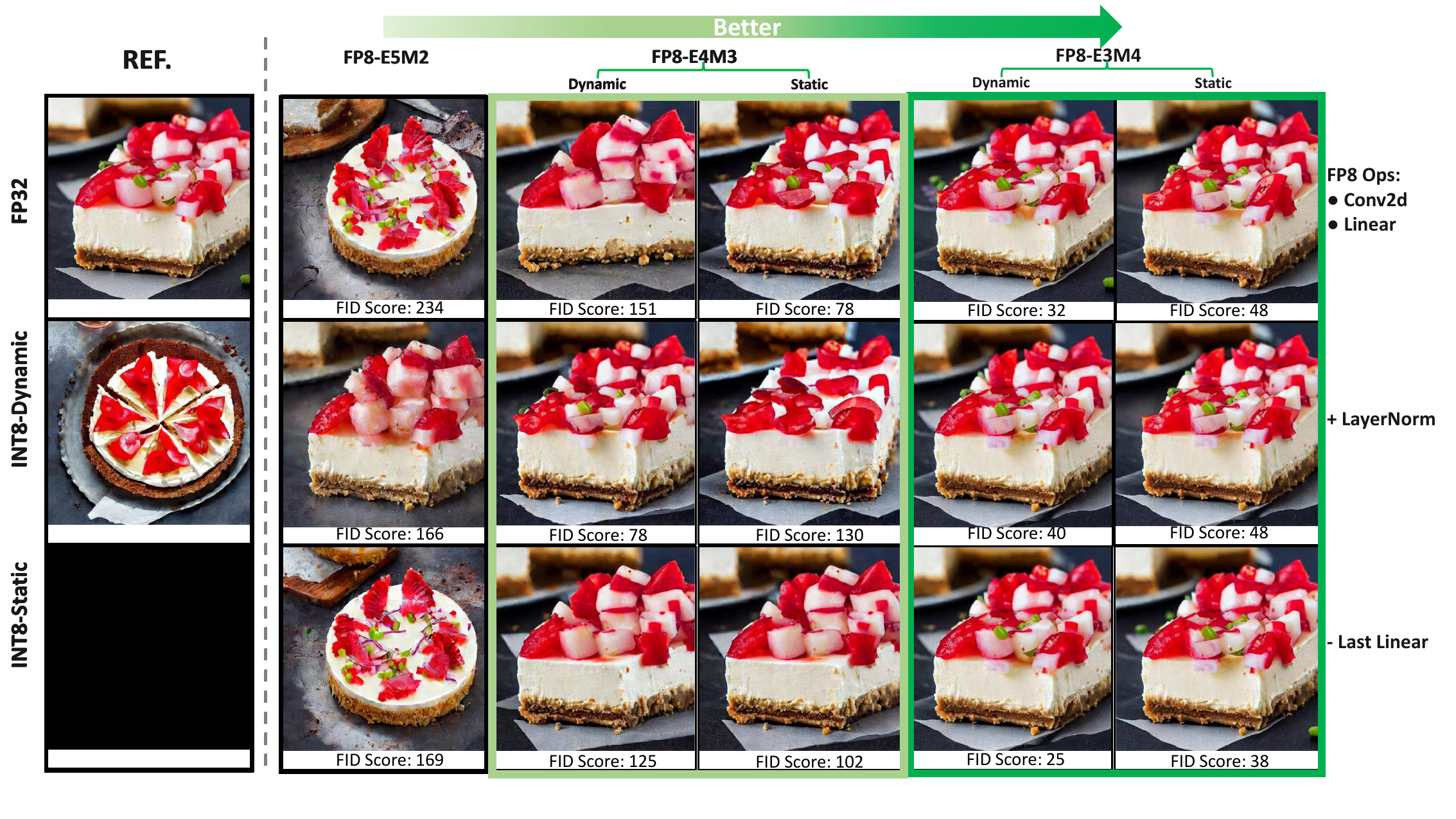}
    \caption{Stable Diffusion with Prompt: "A delicious ceviche cheesecake slice"}
    \label{fig:hfp8_stable_cake}
\end{figure*}

\begin{figure*}[htbp]
    \centering
    \includegraphics[scale=0.45]{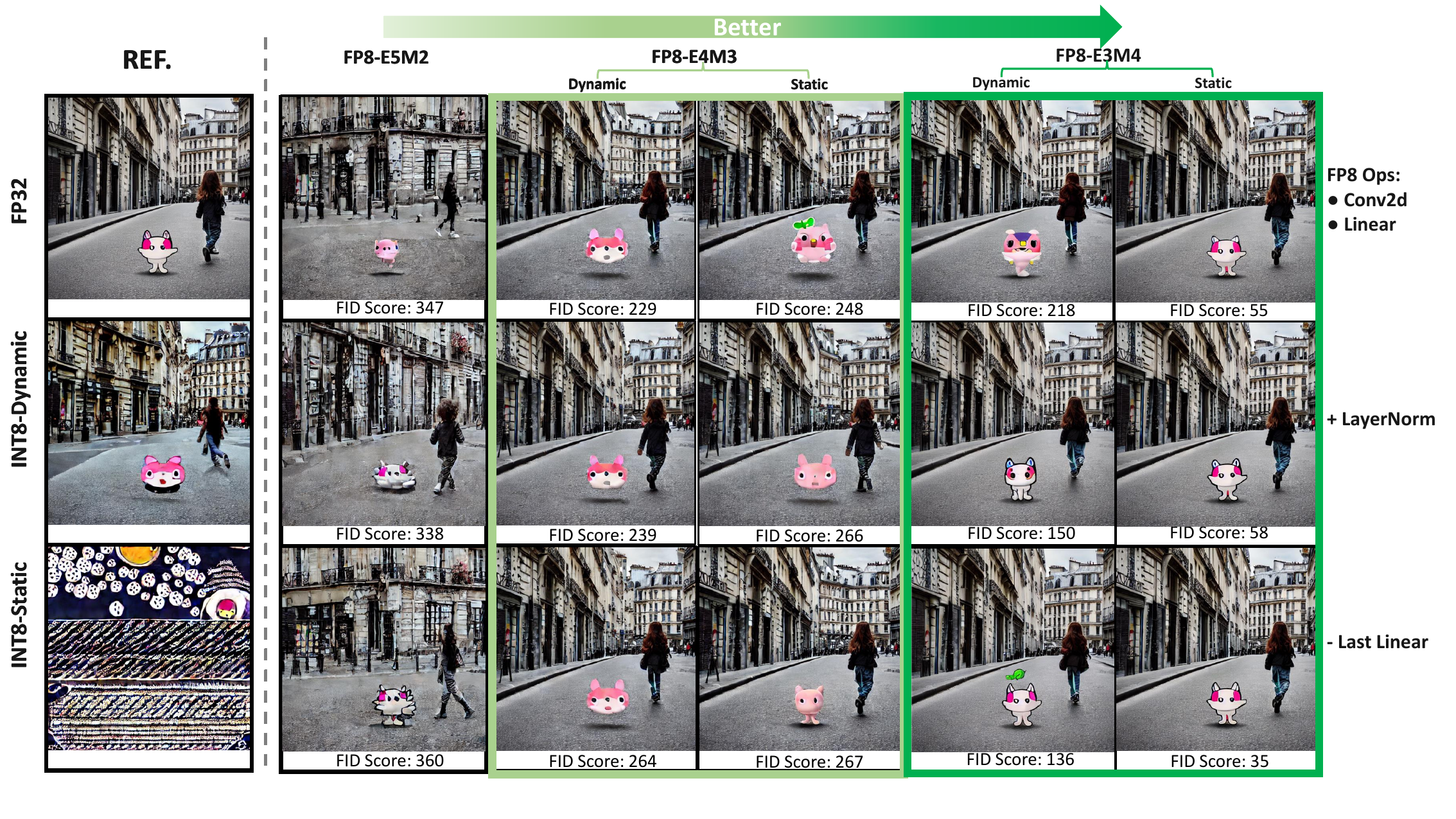}
    \caption{Stable Diffusion with Prompt: "The spirit of a tamagotchi wandering in the city of Paris"}
    \label{fig:hfp8_stable_pokemon}
\end{figure*}

Besides the sample generated with the prompt "A photo of an astronaut riding a horse on Mars", we also generate two another images with different prompts as shown in Figure~\ref{fig:hfp8_stable_cake} and~\ref{fig:hfp8_stable_pokemon}.

\subsection{Text Generation Samples from BLOOM}
Table~\ref{tbl:bloom} shows generated text of BLOOM on different data formats and quantization approaches.
\begin{table*}[ht!]
	\centering
	\caption{Generated Text of BLOOM (32 Input Tokens and 100 Output Tokens)}
        \vskip 0.15in
	\begin{tabularx}{\textwidth}{XXX}
		\toprule
		  \textbf{Prompt}: Once upon a time, there existed a little girl, who liked to have adventures. She wanted to go to places and meet new people, and have fun. \\
		\midrule
            \textbf{Output (FP32)}: One day, she decided to go on a trip. She packed her suitcase and went to the airport. When she got there, she found out that there was no flight to her destination, so she decided to take a bus. When she got there, she found out that there was no bus to her destination, so she decided to take a train. When she got there, she found out that there was no train to her destination, so she decided to take a car. When she got \\
		\midrule
    	\textbf{Output (INT8)}: This little girl was very adventurous. One day she decided to go on a trip to a faraway country. When she got there the little girl saw many strange things. She saw many strange people. She saw many strange animals. She saw many strange sights. She saw many strange smells. She saw many strange sounds. She saw many strange sights. She saw many strange smells. She saw many strange sounds. She saw many strange sights. She saw many strange smells. She saw many \\
		\midrule
            \textbf{Output (E5M2)}: One day, she decided to go on a trip with her parents. They took her to the seaside, where they had a wonderful time. After that, they went on a trip to the mountains, where they had a wonderful time. Then they went on a trip to the city, where they had a wonderful time. After that, they went on a trip to the countryside, where they had a wonderful time. Then they went on a trip to the zoo, where they had a wonderful time \\
		\midrule
            \textbf{Output (E4M3 Dynamic)}: One day, she decided to go on a journey. She packed her suitcase, and set off on her journey. She was very excited about her trip, and she had a lot of things to do. First, she went to the supermarket, and bought a lot of food. Then, she went to the bakery, and bought a lot of bread. Then, she went to the butcher, and bought a lot of meat. Then, she went to the florist, and bought \\
		\midrule
            \textbf{Output (E4M3 Static)}: One day, she decided to go on a trip. She packed her suitcase and went to the airport. When she got there, she found out that her flight was delayed. She waited for a long time, and when she finally got on the plane, she found out that it was going to take a long time to get to her destination. When she finally got to her destination, she found out that her flight had been canceled. She waited for a long time, and when she \\
		\midrule
            \textbf{Output (E3M4 Dynamic)}: One day, she decided to go on a trip with her parents. They took her to the zoo, where she saw a lion, a tiger, a bear, and a monkey. The lion was the biggest, the tiger was the fiercest, the bear was the strongest, and the monkey was the smartest. The lion ate the tiger, the tiger ate the bear, the bear ate the monkey, and the monkey ate the lion. The lion, the tiger, the bear, and  \\
		\midrule
            \textbf{Output (E3M4 Static)}: One day, she decided to go on an adventure. She packed her suitcase and went to the airport. She boarded a plane and flew to New York City. There, she met a man, and they had a great time together. They went to a restaurant and ate delicious food. Then, they went to the park and had a great time together. They went to the zoo and had a great time together. They went to the beach and had a great time together. They went to \\
		\midrule
            \textbf{Output (FP8 Mixed)}: One day, she decided to go on a trip. She packed her suitcase and went to the airport. When she got there, she took a taxi to her hotel. When she got to her room, she unpacked her suitcase and went to the bathroom. She washed her face, brushed her teeth, and put on her makeup. When she was done, she put on her clothes and went to the restaurant. When she got there, she sat down at a table and ordered a \\
		\bottomrule
	\end{tabularx}%
	\label{tbl:bloom}
        \vskip 0.15in
\end{table*}%




\end{document}